\documentclass{article}
\usepackage{spconf,amsmath,graphicx}
\usepackage{multirow}
\usepackage{booktabs}
\usepackage[table,xcdraw]{xcolor}
\usepackage{amssymb}
\usepackage{dsfont}

\title{Learning Multiscale Consistency for Self-supervised Electron Microscopy Instance Segmentation}
\name{Yinda Chen$^{1, 2}$, Wei Huang$^2$, Xiaoyu Liu$^2$, Shiyu Deng$^2$, Qi Chen$^2$, Zhiwei Xiong$^{1, 2}$}
\address{$^{1}$ Institute of Artificial Intelligence, Hefei Comprehensive National Science Center \\
        $^{2}$ University of Science and Technology of China}

\begin{document}
%
\maketitle
\begin{abstract}
Instance segmentation in electron microscopy (EM) volumes is tough due to complex shapes and sparse annotations. Self-supervised learning helps but still struggles with intricate visual patterns in EM.
To address this, we propose a pretraining framework that enhances multiscale consistency in EM volumes. Our approach leverages a Siamese network architecture, integrating both strong and weak data augmentations to effectively extract multiscale features. We uphold voxel-level coherence by reconstructing the original input data from these augmented instances. Furthermore, we incorporate cross-attention mechanisms to facilitate fine-grained feature alignment between these augmentations. Finally, we apply contrastive learning techniques across a feature pyramid, allowing us to distill distinctive representations spanning various scales.
After pretraining on four large-scale EM datasets, our framework significantly improves downstream tasks like neuron and mitochondria segmentation, especially with limited finetuning data. It effectively captures voxel and feature consistency, showing promise for learning transferable representations for EM analysis.
\end{abstract}
\section{Introduction}
\label{sec:intro}

Electron microscopy (EM) imaging plays a vital role in providing high-resolution
views of cellular tissue structures, especially in the nervous system \cite{sheridan2022local}.
With
the development of deep learning techniques, automatic segmentation for EM volumes has
become an indispensable tool for analyzing the huge amount of data \cite{funke2018large,li2022advanced}.
However, existing fully supervised EM instance segmentation methods are fragile due to a lack of sufficient annotations \cite{huang2022semi}.

Self-supervised learning exploits the inherent structure and information within
data to acquire meaningful representations without the supervision of manual
annotations, which brings performance improvement in downstream tasks \cite{he2020momentum,wan2023med,liu2023m}. However, existing self-supervised pretraining methods often rely on simple and generic
proxy tasks, such as image reconstruction \cite{zhou2022self,chenself}, which cannot fully exploit the complex and diverse visual representations in EM volumes.

To address this issue, we propose a multiscale consistency pretraining framework
tailored for EM volumes. The framework focuses on learning visual representation consistency
at different scales through joint training on multiple tasks, including voxel-level reconstruction,
soft feature matching, and feature contrastive learning. To obtain robust EM
visual representations, we utilize a Siamese network architecture with a CNN or
Transformer backbone to extract features from strong and weak augmented
EM volume inputs. To maintain voxel-level consistency in the latent representations, we
first enforce the network to reconstruct the original volume from its
augmented counterparts. To achieve fine-grained feature-level consistency, we then adopt
a cross-attention mechanism for soft feature matching between strong and weak augmentations.
Moreover, we propose a contrastive learning scheme on the feature pyramid,
which utilizes a hierarchical approach to compute similarity scores across different
levels of the pyramid. In this way, we promote the network to learn invariant
features on multiple scales, thereby enhancing its robustness to complex instance morphology in EM volumes.

Our contributions are as follows: 1) We propose a novel framework tailored for EM
volume pretraining, enforcing both voxel-level and feature-level consistency in EM visual representations. 2) We utilize a Siamese network with cross-attention
to focus on features at different scales for fine-grained feature matching. 3) We conduct
extensive pretraining on four large-scale EM datasets using both CNN and
Transformer backbones and demonstrate the effectiveness of our framework on
downstream tasks of neuron segmentation and mitochondria segmentation.

\section{Methodology}
\label{sec:Methodology}
\begin{figure}[t]
    \centering
    \includegraphics[width = \linewidth]{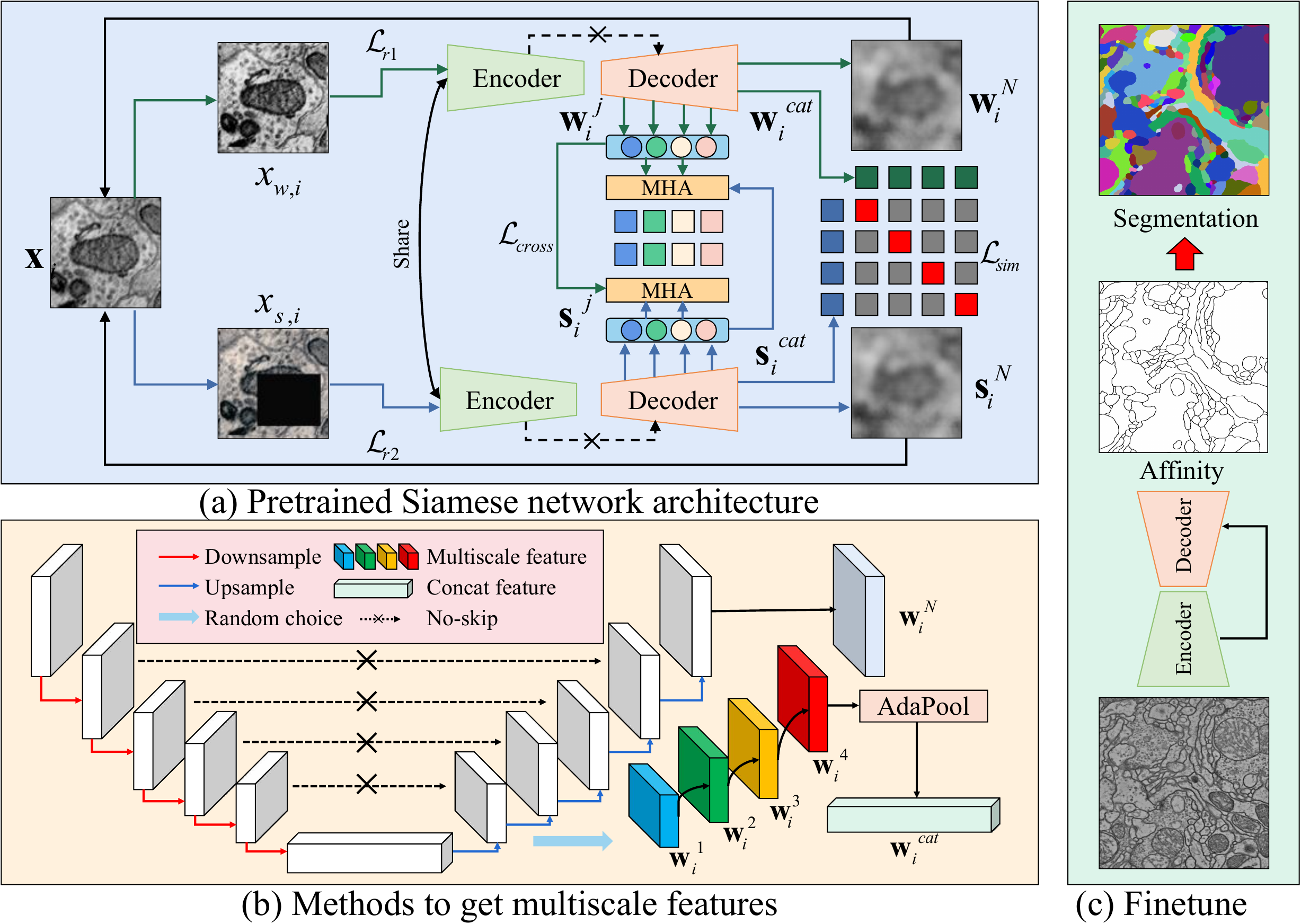}
    \caption{Our proposed method consists of two stages: pretraining and finetuning. (a) illustrates the framework of pretraining, which aims to add consistency constraints to both weak and strong augmented data. (b) describes the method to extract multiscale features. (c) outlines our instance segmentation method during the finetuning stage.}
    \label{fig:pipeline}
\end{figure}
\subsection{Siamese Network Design}
To enhance the robustness of the network and improve its ability to extract features at different scales, we feed both strong augmented and weak augmented data into a Siamese network as shown in Figure \ref{fig:pipeline}(a). Strong augmentation techniques include elastic deformation, Gaussian noise, gamma transformation, cropping, splicing, scaling, and large-scale masking, while weak augmentation techniques consist of simple transformations such as translation, rotation, scaling, and flipping. We transform the original 3D pretraining data $\mathcal{D}$ into $\mathcal{D'}=\{\left(x_{w, 1}, x_{s, 1}\right),\allowbreak\left(x_{w, 2}, x_{s, 2}\right), \ldots,\allowbreak\left(x_{w, N}, x_{s, N}\right)\}$, where ${x_{w, i}, x_{s, i}}$ represent the weak and strong augmented inputs, respectively, with ${x_{w, i}, x_{s, i}} \in \mathbb{R}^{B \times C \times H \times W \times D}$.
Here, $B$ represents the batch size, $C$ represents the number of channels, and $H$, $W$, and $D$ represent the dimensions of the volume. The transformed data are then mapped into the latent space using a shared-weight backbone, such as UNet or ViT \cite{dosovitskiy2020image}, to generate features at different scales. For example, given $x_{w,1}$ as input, the decoder of our main network generates a series of encoded visual features $\mathbf{W}_1 = \{\mathbf{w}_1^1,\mathbf{w}_1^2,....,\mathbf{w}_1^N\}$, where $\mathbf{w}_i^j \in \mathbb{R}^{B \times C_1 \times H_1 \times W_1 \times D_1}$. 
To ensure consistent output sizes, we pass the multiscale features through an Adaptive Average Pooling layer, resulting in encoded features with dimensions of $H_i=16, W_i=16, D_i=16$. Moreover, we eliminate the skip connections in the UNet structure to stop the network from producing degenerate solutions based on input features from the encoder \cite{zhou2023unified}.

\subsection{Pretraining Method}
\subsubsection{Voxel-level Reconstruction.}
Self-supervised reconstruction tasks can be used to learn visual representations with good reconstruction performance \cite{huang2022semi}. By embedding input data into a high-dimensional space and then reconstructing it back to the original input, the model can learn a robust representation of the input data. Therefore, we added a consistency reconstruction objective to the output of the Siamese network, aimed at reconstructing the original input under strong and weak data augmentation. The loss function of the reconstruction task can be formulated as follows:\vspace{-0.2cm}
\begin{equation}
\vspace{-0.2cm}
\mathcal{L}_{{r}} = \frac{1}{2N}\sum_{i=1}^{N}(||\mathbf{w}_i^{N}-\mathbf{x}_i||_2^2+||\mathbf{s}_i^{N}-\mathbf{x}_i||_2^2),
\end{equation}
where $N$ signifies the total number of intermediate output features within the network.

\subsubsection{Soft Feature Matching based on Cross-attention.}
Cross-attention based feature match does not require explicit feature extraction and handles larger variations between input data \cite{wangmulti}. Since our Siamese network introduces complex deformations and perturbations, we introduce a bidirectional cross-attention mechanism to match fine-grained features. The specific technical details are as follows.

For each pair of strong-weak augmentation $(\mathbf{W_i},\mathbf{S_i})$, we first project them into lower-dimensional embeddings using $\allowbreak{\mathbf{W}}^e_i=mlp(\mathbf{w_i^1},\mathbf{w_i^2},....,\mathbf{w_i^N})$ and ${\allowbreak\mathbf{S}}^e_i=mlp(\mathbf{s_i^1},\mathbf{s_i^2},....,\allowbreak\mathbf{s_i^N})$. To enhance the correlation between strong and weak augmentation tokens, we propose to utilize a cross-attention mechanism to compute soft feature matching between generated strong-weak augmentation tokens. Specifically, for the $j$-th strong augmentation token embedding ${\mathbf{w}}_i^j$ in the $i$-th strong-weak augmentation pair, we let ${\mathbf{w}}_i^j$ attend to all weak augmentation token embeddings in ${\mathbf{S}}_i$. We first compute bidirectional cross-attention, $v_i^{j 2 k} = \text{softmax}\left[\frac{\left(Q {\mathbf{w}}_i^j\right)^T\left(K {\mathbf{s}}_i^k\right)}{\sqrt{d}}\right]$,$\allowbreak\omega_i^{j 2 k} = \text{softmax}\left[\frac{\left(Q{\mathbf{s}}_i^j\right)^T\left(K {\mathbf{w}}_i^k\right)}{\sqrt{d}}\right]$,
and then calculate their attention weights $\mathbf{m}_i^j=\sum_{k=1}^N\allowbreak\left[v_i^{j 2 k}\left(V {\mathbf{s}}_i^k\right)\right]$ and $\mathbf{n}_i^j=\sum_{k=1}^N\allowbreak\left[\omega_i^{j 2 k}\left(V {\mathbf{w}}_i^k\right)\right]$, where K, Q, and V are learnable matrices. We introduce InfoNCE loss \cite{he2020momentum} to pull the corresponding strong-weak token embeddings closer but push them away from other token embeddings, and the resulting cross-attention loss is calculated as follows:\vspace{-0.1cm}
\begin{equation}
\resizebox{\linewidth}{!}{$
\mathcal{L}_{\text{cross}} = -\frac{1}{N B} \displaystyle\sum\limits_{i=1}^{N}\displaystyle\sum\limits_{j=1}^{B} \log\frac{e^{\mathbf{m_j^i}^T\cdot \mathbf{n_j^i}/\tau}}{e^{\mathbf{m_j^i}^T\cdot \mathbf{n_j^i}/\tau} + \sum_{k=1, k\neq i}^{N} e^{\mathbf{m_j^i}^T\cdot \mathbf{n_k}/\tau}},$}
\end{equation}
where $\tau$ represents the temperature hyperparameter, we have set it to 0.07.

\vspace{-0.3cm}
\subsubsection{Multiscale Feature Contrastive Learning.}
To further consider the importance of multiscale features in Electron Microscopy (EM) volumes and reduce the computational overhead associated with conventional contrastive learning, inspired by recent work like \cite{zhou2023unified,mo2023multi}, we adopt a 
 approach for Self-Supervised Learning (SSL) that minimizes distances between feature pairs $\mathbf{w}_i^{j}$ and $\mathbf{s}_i^{j}$ at different scales, encouraging the model to retain multiscale self-supervised representations, as shown in Figure \ref{fig:pipeline}(b).

For a given feature pair $(\mathbf{w}_i^j, \mathbf{s}_i^j)$, we begin by randomly selecting a single channel from each feature. Subsequently, we employ adaptive average pooling to standardize the feature sizes, and concatenate the features from different levels, resulting in $\mathbf{w}_i^{cat}$ and $\mathbf{s}_i^{cat}$, and then flatten these features to compute their similarity. To introduce some asymmetry and prevent model collapse, as suggested in \cite{li2023frozen,zbontar2021barlow,chen2023generative}, we add an additional MLP encoding layer to the vector produced by strong augmentation, resulting in output vectors $\mathbf{w}_i^{cat}$ and $mlp(\mathbf{s}_i^{cat})$. Subsequently, we compute the similarity of multiscale features from the Siamese network using cosine similarity. We calculate the multiscale similarity loss $\mathcal{L}_{sim}$ for the entire batch as follows:
\vspace{-0.2cm}
\begin{equation}
\resizebox{\linewidth}{!}{$
\mathcal{L}_{\text{sim}}=\hspace{-0.1cm}\displaystyle\sum\limits_{\substack{i=1}}^B -\frac{1}{2} \left[\cos \left(\left(\mathbf{w}_i^{cat}\right), mlp\left(\mathbf{s}_i^{cat}\right)\right)\right.  \left.\hspace{-0.1cm}+\cos \left(mlp\left(\mathbf{s}_i^{cat}\right), \left(\mathbf{w}_i^{cat}\right)\right)\right],$}
\end{equation}
where cos refers to cosine similarity, $N$ signifies the total number of intermediate output features within the network, and $mlp$ denotes a fully connected neural network.

Our pretraining method's loss function is mainly composed of the three aforementioned parts. Therefore, the final loss function can be described as follows:\vspace{-0.1cm}
\begin{equation}
\mathcal{L}_{{total}} = \alpha_1 \mathcal{L}_r + \alpha_2 \mathcal{L}_{cross} + \alpha_3 \mathcal{L}_{sim},
\end{equation}
where $\alpha_1$, $\alpha_2$, and $\alpha_3$ represent the weighting coefficients with values of 1, 0.1, and 0.1, respectively.

\section{Experiments and Analysis}
\label{sec:pagestyle}
\subsection{Dataset.}
Our pretraining dataset comprises four large-scale electron microscopy (EM) datasets, namely, the Full Adult Fly Brain (FAFB) \cite{schlegel2021automatic}, MitoEM \cite{wei2020mitoem}, FIB-25 \cite{takemura2017connectome}, and Kasthuri15 \cite{kasthuri2015saturated} datasets. These datasets originate from various organisms, including Drosophila, mice, rats, and humans. To ensure a balanced and diverse input for our network, we uniformly sample from these four datasets with equal probabilities for each batch. The total size of our pretraining dataset is approximately 400 GB.

To validate the effectiveness of our pretraining method, we conduct experiments for neuron and mitochondria segmentation tasks in the finetuning stage.
For neuron segmentation, we utilize the CREMI A/B/C datasets \cite{funke2016miccai}, training on 75 or 10 of the images and testing on the remaining 50 images.
For the mitochondria segmentation task, we adopt the MitoEM-R dataset \cite{wei2020mitoem}, training on the first 400 or 100 images and testing on the remaining 100 images.

\subsection{Evaluation Metrics and Hyperparameters.}
In the finetuning stage, we evaluate our pretraining method on two 3D instance segmentation tasks: neuron segmentation and mitochondria segmentation. As illustrated in Figure \ref{fig:pipeline}(c), following \cite{huang2022learning,li2022advanced}, we predict the instance's affinity map or boundary map through the network and generate specific instances via post-processing methods waterz \cite{funke2018large}. We load the weights obtained from pretraining to improve the final instance segmentation performance and demonstrate the effectiveness of our pretraining method.

For neuron instance segmentation, we evaluate performance using the Variation of Information (VOI) \cite{nunez2013machine} and Adapted Rand (Arand) \cite{arganda2015crowdsourcing} metrics, while for mitochondria segmentation, we use the average precision with an IoU threshold of 0.75 (AP75) metric. Additionally, following \cite{wei2020mitoem}, we provide evaluation metrics for different sizes of mitochondria instances based on the number of mitochondria voxels, with thresholds set at 5K and 15K for small, medium, and large instances, respectively.

We employ the Adam optimizer with $\beta_1=0.9, \beta_2=0.999$ and set the learning rate to $1 \times 10^{-4}$ for both pretraining and finetuning. Pretraining is conducted with a batch size of 16 on 8 NVIDIA RTX 3090s GPUs. We randomly sample the four large-scale EM datasets with equal probability and train the network for a total of $500$K iterations. In the finetuning phase, we set the batch size to 4 on 2 NVIDIA RTX 3090s GPUs, and conduct experiments on both neuron and mitochondria segmentation tasks, with $200$K iterations for each task.

\begin{table}[t]
\fontsize{9}{10.6}\selectfont
\centering
\renewcommand\tabcolsep{4.5pt}
\caption{Quantitative comparisons for neuron segmentation on the CREMI datasets.}
\label{tab:neuron}
\begin{tabular}{lcccc}
\toprule[1.2pt]
\multirow{2}{*}{Methods}   & \multicolumn{2}{c}{VOI $\downarrow$} & \multicolumn{2}{c}{ARAND $\downarrow$} \\
                    & 10         & 75         & 10          & 75          \\ \midrule
Random              & 4.432      & 1.497      & 0.771       & 0.236       \\
BYOL \cite{grill2020bootstrap}        & 2.971      & 1.445      & 0.559       & 0.229       \\
SimSiam \cite{chen2021exploring} & 2.544      & 1.413      & 0.467       & 0.213       \\
BarlowTwins \cite{zbontar2021barlow} & 2.605      & 1.415      & 0.546       & \underline{0.186}       \\
PCRLv2 \cite{zhou2023unified}      & \underline{2.496}      & \textbf{1.316}      & \underline{0.460}       & 0.196       \\
Ours                & \textbf{2.369}      & \underline{1.366}      & \textbf{0.449}       & \textbf{0.173}       \\ \bottomrule[1.2pt]
\end{tabular}
\vspace{-0.2cm}
\end{table}

\vspace{-0.2cm}
\subsection{Neuron Segmentation.}
\begin{figure}[t]
    \centering
    \includegraphics[width = \linewidth]{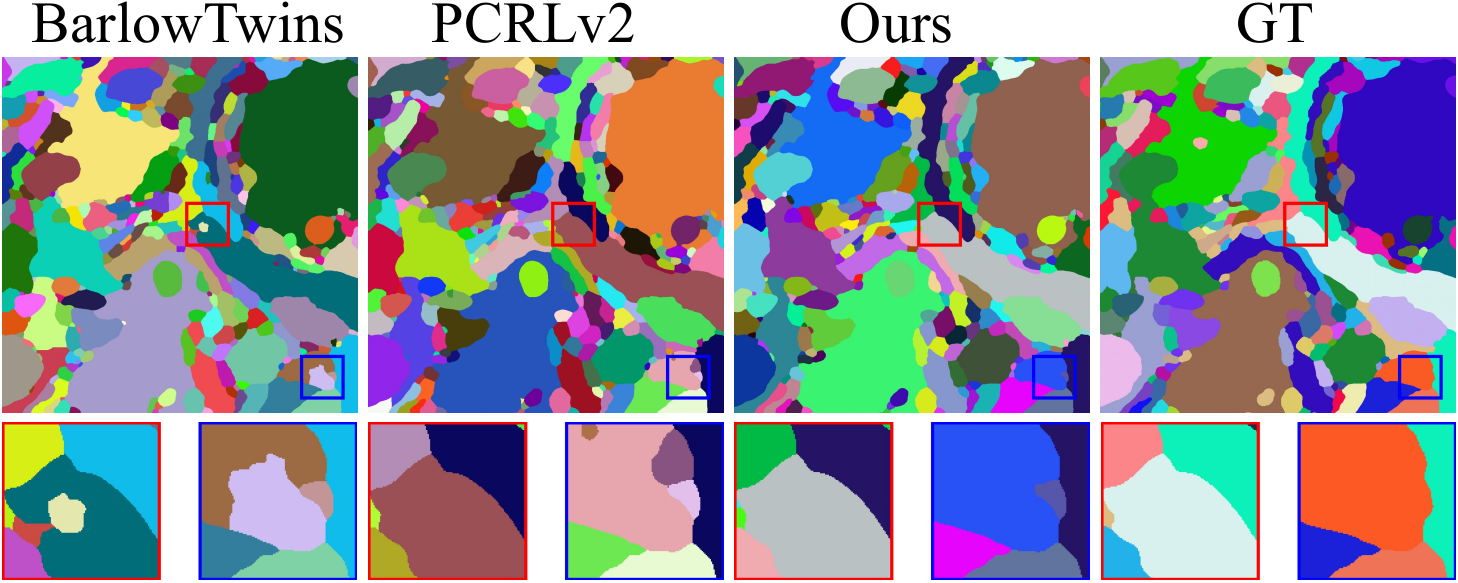}
    \caption{Visual comparisons for neuron segmentation, where colored boxes highlight the significant differences in the segmentation results.}
    \vspace{-0.1cm}
    \label{fig:neuronseg}
\end{figure}

We employ the ViT-base \cite{dosovitskiy2020image} model as an encoder and utilize the UNETR \cite{hatamizadeh2022unetr} approach for neuron segmentation. Additionally, we have reproduced three state-of-the-art self-supervised learning methods commonly used in natural image analysis, namely BYOL \cite{grill2020bootstrap}, SimSiam \cite{chen2021exploring}, and BarlowTwins \cite{zbontar2021barlow}. Moreover, we have incorporated the latest state-of-the-art algorithm designed specifically for medical images, PCRLv2 \cite{zhou2023unified}. The segmentation performance for neuron detection is presented in Table \ref{tab:neuron}, and visualization results are available in Figure \ref{fig:neuronseg}.

Our approach has demonstrated promising performance improvements in the task of neuron segmentation based on ViT backbone networks. Table \ref{tab:neuron} presents the average performance across the CREMI A/B/C datasets. We have observed that our pretraining method excels when fine-tuning with a limited amount of data. For instance, when employing the waterz method for post-processing, the VOI gain averages 8.6\% for fine-tuning with 75 images, while it impressively reaches 46.8\% for fine-tuning with only 10 images. This underscores the exceptional effectiveness of our approach in scenarios with constrained annotation resources.
\vspace{-0.15cm}
\subsection{Mitochondria Segmentation.}
\begin{figure}[tb]
    \centering
    \includegraphics[width = \linewidth]{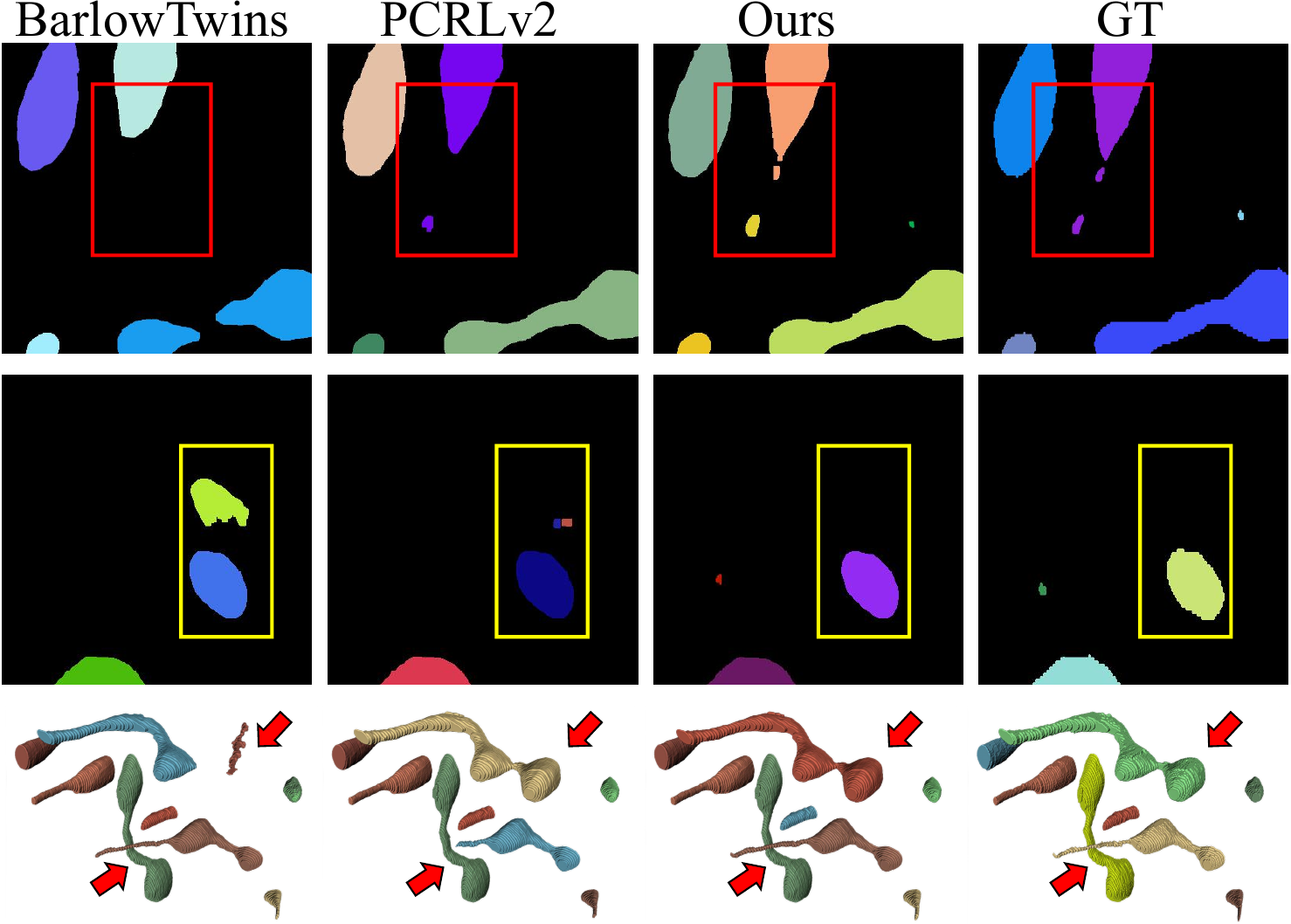}
    \caption{Visual comparisons for mitochondria segmentation, where colored boxes and arrows highlight the significant differences in 2D and 3D segmentation results, respectively.}
    \label{fig:mitoseg}
\end{figure}

For the mitochondria segmentation task, we employed a resnet-50 as the backbone and employed ResUNet \cite{li2022advanced} as the segmentation network, using the same pretraining strategy as for the neuron segmentation task. We fine-tuned these models on the first 400/100 images, respectively, and tested them on the last 100 images. The mitochondria segmentation results are presented in Table \ref{tab:mito}, and detailed visualizations are shown in Figure \ref{fig:mitoseg}.

Our experimental findings demonstrate a significant performance improvement when dealing with challenging instances, specifically in small-volume mitochondria segmentation. This suggests that our pretraining approach substantially enhances segmentation performance on challenging examples. Moreover, in our experimental design, even with the use of only 1/4 of the data for fine-tuning, our method nearly matches the results obtained by fine-tuning on the entire dataset. This underscores the considerable potential of our pretraining approach in enhancing segmentation performance on difficult samples.
\begin{table}[t]
\fontsize{9}{10.6}\selectfont
\centering
\renewcommand\tabcolsep{3pt}
\caption{Quantitative comparisons for mitochondria segmentation on the MitoEM-R dataset.}
\label{tab:mito}
\begin{tabular}{lcccccc}
\toprule[1.2pt]
\multirow{2}{*}{Methods} & \multicolumn{3}{c}{400}                          & \multicolumn{3}{c}{100}                          \\
                         & small          & medium         & large          & small          & medium         & large          \\ \midrule
Random                   & 0.355          & 0.821          & 0.914          & 0.192          & 0.776          & 0.793          \\
BYOL \cite{grill2020bootstrap}             & 0.365          & 0.825          & 0.903          & 0.212          & 0.795          & 0.783          \\
SimSiam \cite{chen2021exploring}          & 0.361          & 0.819          & 0.917          & 0.223          & 0.799          & 0.816          \\
BarlowTwins \cite{zbontar2021barlow}      & 0.377          & 0.831          & 0.923          & 0.209          & 0.814          & 0.827          \\
PCRLv2 \cite{zhou2023unified}           & \underline{0.381}          & \underline{0.853}          & \textbf{0.939} & \underline{0.329}          & \underline{0.821}          & \underline{0.856}          \\
Ours                     & \textbf{0.392} & \textbf{0.859} & \underline{0.931}          & \textbf{0.357} & \textbf{0.842} & \textbf{0.881} \\ \bottomrule[1.2pt]
\end{tabular}
\end{table}
\vspace{-0.15cm}
\subsection{Ablation Study.}

We present ablation experimental results regarding the multi-task constraints in Table \ref{tab:ablation1}. Specifically, we fine-tune our model using the first 75 images from the CREMI C dataset with a ViT-base backbone and observe a maximum performance improvement of up to 34.3\% compared to the baseline method. Additionally, we note that the soft feature matching within the cross-attention mechanism and the reconstruction task exhibit more significant improvements in downstream segmentation performance. 
\begin{table}[tb]\fontsize{9}{10.6}\selectfont
\centering
\renewcommand\tabcolsep{2pt}
\vspace{-0.3cm}
\caption{Ablation study on the multi-task pretraining.}

\begin{tabular}{c|ccc|cc}
\toprule[1.2pt]
Dataset                 & $L_{sim}$ & $L_{r}$ & $L_{cross}$ & VOI $\downarrow$   & Arand $\downarrow$ \\ \midrule
\multirow{5}{*}{CREMI C} &               &                &             & 1.924         & 0.192         \\
                        &               &                & \checkmark     & 1.779(+0.145)  & 0.204(-0.012)  \\
                        &               & \checkmark      &             & 1.601(+0.323) & 0.178(+0.014) \\
                        &               & \checkmark        & \checkmark     & 1.444(+0.480) & \textbf{0.153(+0.039)} \\
                        & \checkmark       & \checkmark        & \checkmark     & \textbf{1.433(+0.491)} & 0.162(+0.030) \\ \bottomrule[1.2pt]
\end{tabular}
\vspace{-0.2cm}
\label{tab:ablation1}
\end{table}
\vspace{-0.15cm}
\section{Conclusion}
In this paper, we introduce a novel pretraining framework for EM volume analysis that leverages voxel-level reconstruction, soft feature matching based on cross-attention, and multiscale feature contrastive learning. By imposing consistent output constraints on the strong-weak data augmentation inputs of a Siamese network, our framework maintains both voxel-level and feature-level consistency in the latent representations of EM volumes. We conduct pretraining on four large-scale EM datasets to ensure data diversity and balance. Compared to existing self-supervised frameworks, our approach demonstrates promising performance improvements in neuron and mitochondria instance segmentation, particularly in cases with complex instance morphology and limited annotations. Moreover, our framework is adaptable to both CNN and Transformer network architectures.  

\vfill\pagebreak

\small
\bibliographystyle{IEEEbib}

\end{document}